\documentclass[letterpaper, 10 pt, conference]{ieeeconf}  

\IEEEoverridecommandlockouts                              

\PassOptionsToPackage{unicode,colorlinks,linkcolor=blue,citecolor=blue,urlcolor=blue}{hyperref}

\usepackage{graphicx} 

\usepackage{amsthm}
\usepackage{pifont}  
\usepackage{booktabs} 
\usepackage{cite}
\usepackage{xcolor}
\usepackage{preamble}
\usepackage{cleveref}
\usepackage{hyperref}

\usepackage{tabularx}
\usepackage{graphicx}
\usepackage{siunitx}
\usepackage{array}
\usepackage{multirow}
\usepackage{threeparttable}
\usepackage[font=footnotesize,labelfont=bf]{caption}
\usepackage[most]{tcolorbox}

\newcolumntype{Y}{>{\centering\arraybackslash}X}

\newcommand{\dt}{\Delta t}

\newcommand{\state}{x}
\newcommand{\yaw}{\theta}
\newcommand{\xref}{{x}^{\mathrm{ref}}}

\newcommand{\uref}{u^{\mathrm{ref}}}
\newcolumntype{P}[1]{>{\centering\arraybackslash}p{#1}} 

\newcommand{\second}[1]{\underline{#1}}
\newcommand{\besttt}[1]{\textbf{#1}}

\makeatletter
\renewcommand\paragraph{\@startsection{paragraph}{4}{0pt}%
  {1ex \@plus 1ex \@minus .2ex}%
  {-1em}%
  {\normalfont\normalsize\itshape}}
\makeatother

\title{\bf
Off-Road Navigation via Implicit Neural Representation\\%
of Terrain Traversability 
}

\author{Yixuan Jia\textsuperscript{1}, Qingyuan Li\textsuperscript{1}, Jonathan P. How\textsuperscript{1}
\thanks{This work was supported in part by Army’s Research
Lab (ARL) under Grant W911NF-21-2-0150 and ARL DCIST under Cooperative Agreement Number W911NF-17-2-0181. Distribution Statement A: Approved for public release; distribution is unlimited.}
\thanks{\textsuperscript{1}Massachusetts Institute of Technology, Cambridge, MA 02139, USA. \{\texttt{yixuany, andyli27, jhow}\}\texttt{@mit.edu}.}%
}

\begin{document}

\maketitle
\thispagestyle{empty}
\pagestyle{empty}

\begin{abstract}
Autonomous off-road navigation requires robots to estimate terrain traversability from onboard sensors and plan motion accordingly. 
Conventional approaches typically rely on sampling-based planners such as MPPI to generate short-term control actions that aim to minimize traversal time and risk measures derived from the traversability estimates. These planners can react quickly but optimize only over a short look-ahead window, limiting their ability to reason about the full path geometry, which is important for navigating in challenging off-road environments. Moreover, they lack the ability to adjust speed based on the terrain-induced vibrations, which is important for smooth navigation on challenging terrains.
In this paper, we introduce TRAIL (\underline{Tra}versability with an \underline{I}mplicit \underline{L}earned Representation), an off-road navigation framework that leverages an implicit neural representation to model terrain properties as a continuous field that can be queried at arbitrary locations. This representation yields spatial gradients that enable integration with a novel gradient-based trajectory optimization method that adapts the path geometry and speed profile based on terrain traversability.
\end{abstract}


\section{Introduction}

Autonomous navigation in off-road environments requires capabilities beyond obstacle avoidance. Unlike structured roads, off-road terrain is unstructured and highly variable (e.g., uneven surfaces and vegetation), motivating substantial recent progress in traversability-aware perception and planning \cite{fan2021step,jian2022putn,xue2023traversability,dixit2024step,frey2024roadrunner,patel2024roadrunner,yoo2024traversability,lee2025trg,frey2023fast,cai2023probabilistic,cai2024evora,cai2025pietra,mattamala2025wild,gasparino2022wayfast,gasparino2024wayfaster,maturana2017real,guan2022ga,shaban2022semantic,meng2023terrainnet,castro2022does}.
%

Most existing off-road navigation stacks derive a grid-based traversability cost map from either exteroceptive terrain cues (e.g., geometry or appearance) or proprioceptive interaction signals (e.g., slip or vibration). Each cue alone has limitations: geometry or appearance-only estimates can be ambiguous in unstructured terrain (e.g., drivable vegetation versus rigid obstacles), while interaction-based supervision is inherently sparse because it is only observed along the vehicle's traversed trajectories. On the planning side, many systems execute short-horizon, sampling-based control (e.g., MPPI \cite{williams2017information}) over these costs; while reactive and flexible, such approaches have limited ability to refine full-path geometry and to adapt a speed profile to terrain-induced disturbances.

In this work, we address limitations in both perception and planning by combining complementary terrain cues. We estimate (i) geometric properties that provide a conservative yet physically grounded prior and (ii) terrain-response signals that reflect the vehicle's experienced disturbance. We represent these quantities with an implicit neural representation (INR)---a continuous field that can be queried at arbitrary locations---which provides spatial gradients enabling a gradient-based trajectory optimization method that jointly refines path geometry and speed profile subject to various constraints.
In summary, our contributions are:
\begin{itemize}
    \item A learning-based implicit neural representation that models terrain geometry and terrain-response signals as continuous, queryable functions, with readily accessible spatial gradients through lightweight decoders.
    \item A gradient-based trajectory optimization framework that exploits these gradients to co-optimize path geometry and speed profile according to terrain traversability, while naturally handling constraints.
    \item Comprehensive evaluations in simulation and challenging real-world off-road experiments demonstrating improved navigation performance over conventional frameworks.
\end{itemize}

\begin{figure}[!t]
    \centering
    \includegraphics[trim=0 50 0 0, clip, width=0.48\textwidth]{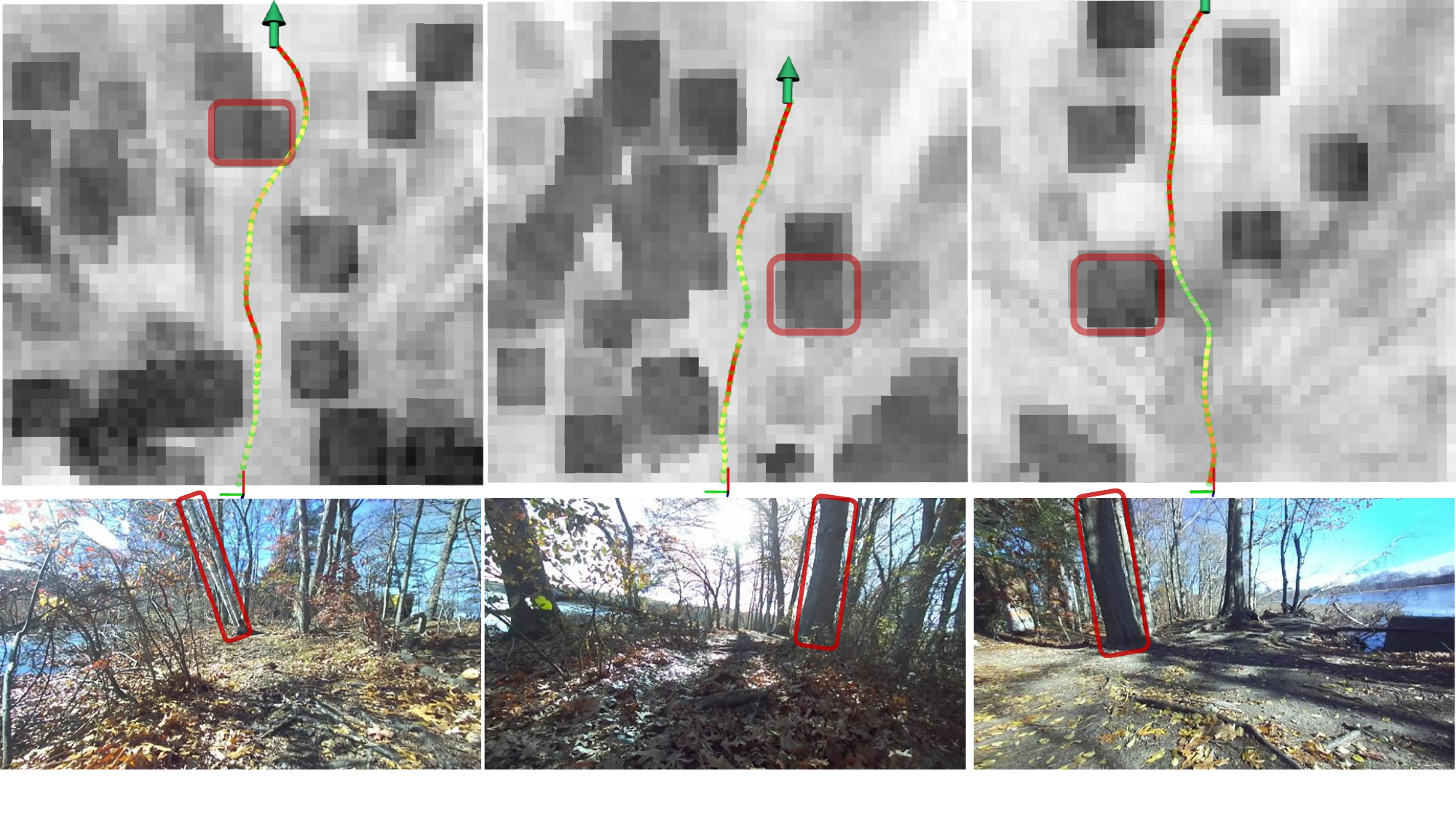}
    \caption{{\bf Top}: Optimized trajectories overlaid on the blended cost map, generated by combining two cost maps with $50\%$ transparency (sampled at fixed grid resolution for visualization). The larger dark blocks represent higher geometric risk inflated using vehicle radius (e.g. the highlighted tree trunks) while finer greyscale variations correspond to predicted terrain bumpiness (e.g. grey regions around trees from tree roots).
    Redder trajectory segments indicate higher speed.
    {\bf Bottom}: Corresponding onboard camera images. 
    The optimized trajectories avoid hard obstacles, slow down when approaching bumpy regions, and speed up in smoother areas.}
    \label{fig:hardwarre_slow_down_demo}
\end{figure}

{
\section{Related Work}
\subsection{Traversability Estimation} 
Most existing off-road navigation frameworks estimate terrain traversability based on one of the three approaches: 

1) Geometry-based and heuristic methods \cite{fan2021step,dixit2024step,frey2024roadrunner,patel2024roadrunner,yoo2024traversability,jian2022putn,xue2023traversability}: 
These methods estimate traversability from geometric/heuristic cues (e.g., slope, roughness, curvature) and sometimes observability cues (e.g., point sparsity)---often derived from LiDAR---either via handcrafted costs or learned predictors trained with stack-generated pseudo-labels (e.g., RoadRunner \cite{frey2024roadrunner,patel2024roadrunner}).
For example, PUTN \cite{jian2022putn} computes terrain slope, flatness, and point cloud sparsity and fuses them into a single cost for downstream planning. 
Many works in this category rely mainly on geometric information of the terrain, which may result in conservative traversability estimates (e.g. tall grass may receive a similar cost to rocks).

2) Appearance-based methods
\cite{maturana2017real,guan2022ga,shaban2022semantic,meng2023terrainnet}:
These methods infer traversability from semantic or appearance cues via terrain classification or segmentation, and then the predicted classes can be mapped to planning costs. 
For example, Shaban et al. \cite{shaban2022semantic} train a neural network to classify LiDAR point clouds into discrete traversability classes, which are subsequently converted into a cost map for planning. 
Works in this category can require substantial manual labeling and may be limited by discrete class definitions.  

3) Interaction-based methods
\cite{frey2023fast,gasparino2024wayfaster,cai2023probabilistic,mattamala2025wild,cai2024evora,cai2025pietra,gasparino2022wayfast,castro2022does}:
These methods estimate terrain traversability using supervision derived from vehicle–terrain interaction signals (e.g., IMU vibration, slip, suspension). 
For example, WayFASTER \cite{gasparino2024wayfaster} defines traction as the ratio of achieved speed to commanded speed and trains a neural network to map sensor observations to a grid map of traction values. 
However, a limitation of these methods is that labels are only available along the vehicle’s traversed trajectories, leading to sparse supervision signals and potentially limited generalization. To improve this, recent works \cite{frey2023fast,triest2024velociraptor,mattamala2025wild,sivaprakasam2025salon} leverage vision foundation models (VFMs), such as DINOv2 \cite{oquab2023dinov2}, to obtain general-purpose visual features first and learn mappings from these features (or some compressed version of these features \cite{sivaprakasam2025salon}) to traversability estimates.

\noindent
\textbf{Relation to this work:}
In this work, we learn an INR that jointly models terrain geometry and terrain-response signals. The resulting continuous, queryable field provides spatial gradients that our planner exploits for trajectory optimization.

\subsection{{Planning}}
UGV planning has been extensively studied in both indoor and outdoor environments \cite{ferguson2007field,krusi2017driving,kurenkov2022nfomp,camps2022learning,xu2023efficient,li2025seb}. 
Common approaches include graph- and sampling-based planners (e.g., A$^\ast$/D$^\ast$, RRT$^\ast$, and state lattices) \cite{krusi2017driving,jian2022putn,xu2023efficient,li2025seb,ferguson2007field,hedegaard2021discrete}, as well as trajectory optimization with differentiable map representations (e.g., neural/implicit fields) \cite{kurenkov2022nfomp,camps2022learning}. 
In more challenging unstructured settings, planning is often coupled with risk-aware objectives; STEP \cite{fan2021step,dixit2024step} formulates planning as risk-aware trajectory optimization solved as a QP and warm-started by an offline trajectory library. 

More recently, many off-road systems execute learned traversability costs through sampling-based MPC, most commonly MPPI \cite{williams2017information}, to generate short-horizon actions \cite{gasparino2022wayfast,gasparino2024wayfaster,cai2024evora,sivaprakasam2025salon}. Such methods are flexible but can be sensitive to objective weighting and may exhibit oscillations when trading off competing terms \cite{wang2025genie}.

\noindent
\textbf{Relation to this work:}
In contrast to MPPI-style short-horizon control, our method uses spatial gradients from the INR to perform gradient-based trajectory optimization that refines an entire local path and co-optimizes a traversability-aware speed profile. Compared with STEP, which warm-starts a QP using an offline trajectory library, we generate an initial path online with A$^\ast$ and refine it online within our planning cycle.



}


\section{Proposed Method}
\begin{figure}[!t]
    \centering
    \includegraphics[width=0.48\textwidth]{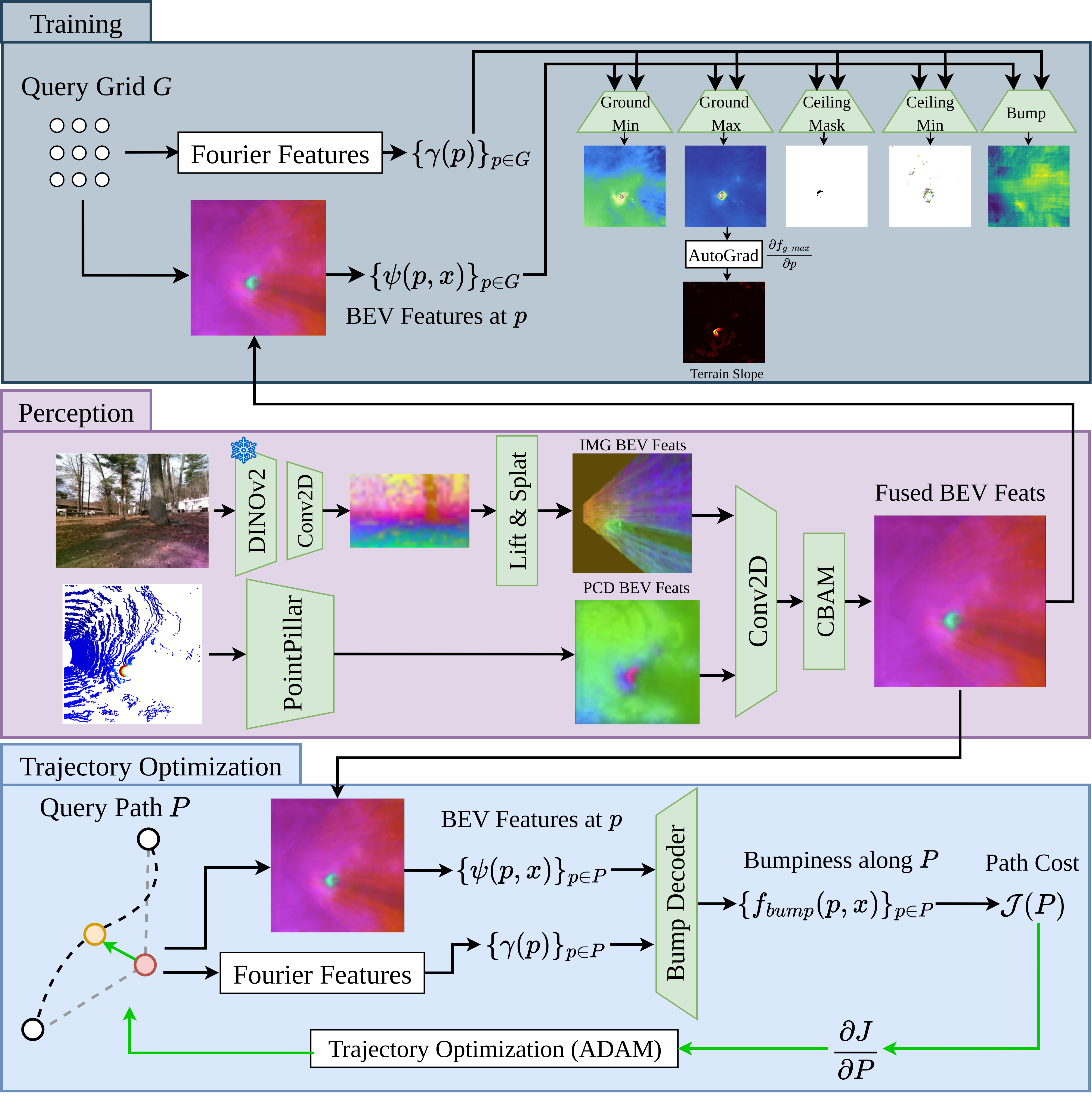}
    \caption{ Our perception module takes a camera image and a LiDAR point cloud and produces an implicit representation of various terrain properties. {During training, we use PyTorch's autograd to obtain the terrain slope directly from ground max elevation $f_{\texttt{g\_max}}$, which provides an extra supervision signal.} During deployment, the proposed planning and control framework incorporates gradient information provided by the implicit representation into a novel gradient-based trajectory optimization framework, where speed profile and path geometry are co-optimized. 
    }
    \label{fig:system_diagram}
    \vspace{-0.5cm}
\end{figure}

Given an RGB image and a LiDAR point cloud from onboard sensors, the task is to infer continuous terrain attributes {(e.g., elevation and slope)} over a local region $[x_{\min},x_{\max}]\times[y_{\min},y_{\max}]$ in the vehicle frame and use them to navigate to goals provided by some higher-level planner (which we assume to be given).


\subsection{{Learning Implicit Representations}}
We train a DNN that maps images and point clouds to traversability-relevant terrain properties. A shared encoder fuses inputs into a BEV feature plane, and query-based decoders output continuous terrain attributes at arbitrary locations (Fig.~\ref{fig:system_diagram}). Unlike fixed-grid predictors \cite{gasparino2024wayfaster,cai2024evora,frey2024roadrunner}, our implicit formulation uses lightweight decoders and provides spatial gradients for trajectory optimization.


\subsubsection{{Encoder}}
Our encoder combines Lift-Splat-Shoot (LSS) \cite{philion2020lift} to project rich image semantics into BEV with PointPillars \cite{lang2019pointpillars} to encode LiDAR-derived 3D structure in BEV, leveraging complementary semantic and geometric cues for more robust traversability prediction.
Given an image $X \in \mathbb{R}^{C \times H \times W}$, we define $\texttt{d}$ as the desired downsampling factor. We extract visual features using DINOv2 \cite{oquab2023dinov2}, obtaining $X_{\texttt{dino}} \in \mathbb{R}^{C_{\texttt{dino}} \times H_p \times W_p}$. These features are compressed with a lightweight 2D convolution block ($\operatorname{Conv} + \operatorname{GroupNorm} + \operatorname{ReLU}$) to produce $X_{\texttt{dc}} \in \mathbb{R}^{C_{\texttt{dc}} \times H_p \times W_p}$.
From $X_{\texttt{dc}}$, we predict depth logits $X_{\texttt{depth}} \in \mathbb{R}^{D \times H_d \times W_d}$, where $H_d = H/\texttt{d}$ and $W_d = W/\texttt{d}$. We also interpolate $X_{\texttt{dc}}$ to obtain 2D image features $X_{\texttt{feat}} \in \mathbb{R}^{C_{\texttt{dc}} \times H_d \times W_d}$.
The pair $(X_{\texttt{feat}}, X_{\texttt{depth}})$ is lifted to a 3D volumetric feature of size $C_{\texttt{dc}} \times D \times H_d \times W_d$ via depth unprojection, which is then splatted onto the BEV plane, following LSS. The resulting image BEV features are compressed to $X_{\texttt{img}} \in \mathbb{R}^{C_{\texttt{img}} \times Y \times X}$.
For LiDAR, we apply PointPillars to obtain point cloud BEV features $X_{\texttt{pcd}} \in \mathbb{R}^{C_{\texttt{pcd}} \times Y \times X}$. The image and point cloud BEV features are concatenated and passed through a 2D convolution block to produce $C_{\texttt{bev}}$ channels, followed by a CBAM attention module \cite{woo2018cbam}. The final encoder output is a fused BEV feature map $X_{\texttt{bev}} \in \R^{C_{\texttt{bev}} \times Y \times X}$.

{
\subsubsection{{Decoder}}
Let the encoder inputs be $\mathbf{X}$ (an image and a point cloud) and let the encoder output a BEV feature map $X_{\texttt{bev}}\in\R^{C_{\texttt{bev}} \times Y \times X}$.
Given a continuous 2D query point $p=(p_x,p_y)$ in the local map frame, we extract a feature vector
$\psi(p,\mathbf{X})\in\mathbb{R}^{C_{\texttt{bev}}}$ by bilinearly sampling $X_{\texttt{bev}}$ at $(p_x,p_y)$.\footnote{We use differentiable bilinear interpolation (e.g., \texttt{grid\_sample}), so $\psi(p,\mathbf{X})$ is differentiable w.r.t.\ $p$.}
Because $X_{\texttt{bev}}$ is produced by convolutional blocks with attention (CBAM), $\psi(p,\mathbf{X})$ aggregates spatial context from a neighborhood around $p$.

We represent each terrain attribute as a coordinate-conditioned MLP (an implicit/coordinate-based decoder) that takes both the local BEV feature and a positional encoding of $p$ as input \cite{peng2020convolutional}.
Directly feeding $p$ to an MLP can underfit high-frequency structure \cite{tancik2020fourier}; we therefore use Fourier features (positional encoding) $\gamma(p)\in\mathbb{R}^{\texttt{ff\_dim}}$:
\begin{equation}
\gamma(p)=\big[\sin(2\pi Bp),\ \cos(2\pi Bp)\big],
\label{eq:fourier_features}
\end{equation}
where $B\in\mathbb{R}^{(\texttt{ff\_dim}/2)\times 2}$ is a fixed random matrix (as in \cite{tancik2020fourier}).
This encoding improves the decoder's ability to represent fine-scale terrain variations (e.g., small rocks or partially buried roots) that matter for traversability.

For compactness, we write the full decoding pipeline as a single function $f_\theta:\mathbb{R}^2\rightarrow\mathbb{R}$:
\begin{equation}
f_\theta(p)=\mathrm{MLP}_\theta\big([\gamma(p);\ \psi(p,\mathbf{X})]\big),
\label{eq:decoder_def}
\end{equation}
which is continuous and differentiable w.r.t.\ $p$. This allows us to query terrain attributes at arbitrary locations and to obtain spatial gradients via automatic differentiation when needed by training and trajectory optimization.

We instantiate five decoders:
$f_{\texttt{g\_min}}$, $f_{\texttt{g\_max}}$, $f_{\texttt{c\_mask}}$, $f_{\texttt{c\_min}}$, $f_{\texttt{bump}}$.
$f_{\texttt{g\_min}}$ and $f_{\texttt{g\_max}}$ predict minimum/maximum ground elevation; $f_{\texttt{c\_mask}}$ and $f_{\texttt{c\_min}}$ predict a ceiling mask and minimum ceiling height (used only during training, inspired by TerrainNet \cite{meng2023terrainnet}, to prevent overhanging structures from inflating ground estimates); and $f_{\texttt{bump}}$ predicts terrain bumpiness as defined in Sec.~\ref{sec:training_data_generation}.
}

\subsubsection{{Data Preparation and Training}}\label{sec:training_data_generation}
The ground truth labels for ground elevations and ceiling predictions are generated in similar fashion to TerrainNet \cite{meng2023terrainnet}, where LiDAR scans in the time range $[t - \delta t, t + \delta t]$ are aggregated using odometry to generate dense annotations at time $t$. Suppose, at a fixed time stamp, each grid has points $\{(x_i, y_i, z_i)\}_{i=1}^k$, ordered by ascending order of $z$. The minimum ground elevation of each grid is then $\frac{1}{m}\sum^m_{i=1}z_i$, where $m$ is a parameter. An average over $m$ points is used instead of $z_1$ in order to smooth out noise. The minimum ceiling elevation is computed by detecting gaps in $z$: if $z_{i+1} - z_i > \delta z$ and $z_{j+1} - z_j \le \delta z ~\forall j < i$, then we classify all points $j$ with $j \ge i+1$ as part of the ceiling, and the ceiling elevation is set to $z_{i+1}$. Once the ceiling is determined, the maximum ground elevation is simply set to $z_i$. In cases where no ceiling is detected, the maximum ground elevation is set to $z_k$. 

To quantify terrain bumpiness, we use the root-mean-square (RMS) of the vehicle's vertical acceleration $a_z$ over a short window (\SI{0.1}{\second}). 
Because the RMS of $a_z$ on the same terrain increases approximately proportionally with vehicle speed \cite{pereira2009robot}, we normalize the measured RMS by the vehicle speed $v$ during dataset generation.
Specifically, for each time $t$ we compute
$\mathrm{rms}_z(t)=\sqrt{\frac{1}{T}\int_{t-T}^{t} a_z(\tau)^2\,d\tau}$ (with $T=0.1$\,s), form the speed-normalized value $\tilde b(t)=\mathrm{rms}_z(t)/(v(t)+\varepsilon)$, and rescale it to $(0,1)$ via $b(t)=\operatorname{sigmoid}(\tilde b(t))$.
We use $b(t)$ as the bumpiness label (i.e. label for $f_{\texttt{bump}}$) and associate it to the corresponding map cell using odometry.
In our trajectory optimization framework (Sec.~\ref{sec:local_planner}), we intentionally exploit the underlying proportional relationship between vibration and speed to encourage speed adaptation based on predicted bumpiness.

To balance the bumpiness data distribution, we compute a histogram of the bumpiness values and give each bumpiness label a weight that is inversely proportional to its frequency.

Finally, for training, we leverage the differentiability of $f_{\texttt{g\_max}}$ to provide extra supervision signal using terrain slope. 
We use PyTorch's autograd functionality to compute the gradient of $f_{\texttt{g\_max}}$ with respect to the query point, $\partial_p f_{\texttt{g\_max}}$, and compute another loss term by comparing $\partial_p f_{\texttt{g\_max}}$ to the ground truth terrain slope (see \Cref{fig:system_diagram}). $L_1$ loss is used for $f_{\texttt{g\_min}},f_{\texttt{g\_max}}, f_{\texttt{c\_min}}, f_{\texttt{bump}}$ and $\partial_p f_{\texttt{g\_max}}$. Cross entropy loss is used for both ceiling mask and depth logits prediction.



\subsection{{Planning with Implicit Representations}}
Our stack combines a geometric path planner with a gradient-based trajectory optimizer that uses $f_{\texttt{bump}}$ to refine the path and speed profile for smoother motion.

\subsubsection{{Path Planner}}
We use A$^\ast$ \cite{hart1968formal} to generate an initial path. To construct the A$^\ast$ cost map, we first query $f_{\texttt{g\_max}}$ at grid locations sampled at a specified coarse resolution. This efficiently produces a coarse geometric grid, from which we compute two metrics: slope (the spatial gradient of $f_{\texttt{g\_max}}$) and step size (the elevation difference between adjacent cells). 
These metrics are normalized to $[0, 1]$ using the maximum acceptable slope and step size respectively.
We take the product of the two metrics to fuse them into a single cost map.
In our A$^\ast$ implementation, the traversal cost of an edge between neighboring cells is given by the average cell cost multiplied by the Euclidean distance between cell centers. The heuristic is chosen as the straight-line distance to the goal multiplied by the minimum cell cost in the map. Because any feasible path must be at least as long as the Euclidean distance and cannot have a per-meter cost lower than this minimum, the heuristic never overestimates the true cost-to-go and is therefore admissible, while still providing a strong guidance signal for the search.


\subsubsection{{Trajectory Optimization}}\label{sec:local_planner}
Given an initial path $\{p^{\text{init}}_i=(p^{\text{init}}_{i, x},p^{\text{init}}_{i, y})\}_{i=1}^{N_p}$ in the local map frame, obtained from A$^\ast$, we refine it via gradient–based trajectory optimization to minimize travel time and vertical acceleration while ensuring smoothness and dynamic feasibility.
We parameterize the geometric path by a set of Catmull–Rom \cite{catmull1974class} control points
$\mathcal{C}=\{c_m\}_{m=1}^{M}$ with fixed endpoints $c_1=p^{\text{init}}_1$, $c_M=p^{\text{init}}_{N_p}$ and trainable midpoints $\{c_2,\dots,c_{M-1}\}$. 
Centripetal Catmull–Rom interpolation~\cite{yuksel2009parameterization} yields a dense path
$\mathbf{P}=\{p_i\}_{i=1}^{N}$ ($N > N_p$) sampled along arc length.

From $\mathbf{P}$ we compute segment lengths $\Delta s_i=\|p_{i+1}-p_i\|$ and path curvature $\kappa_i$. 
We evaluate the bumpiness at each path sample \(p_i\) by aggregating predictions over a
square footprint with side length \(h\), centered at \(p_i\) and aligned with the local yaw. 
This is to account for the vehicle size. Note that each bumpiness prediction is rescaled to be within $(0, 1)$ using $\operatorname{sigmoid}$.
The bumpiness prediction $b_i$ at location $p_i$ is then computed by taking the average bumpiness value over the footprint. For each segment $i$, we then obtain \(\bar b_i=\tfrac{1}{2}(b_i+b_{i+1})\) for \(i=1,\dots,N-1\). 

We also define a curvature-limited speed cap for each segment $i$ as the smooth minimum \cite{nesterov2005smooth} of two speed caps:
\begin{align}
v_{\text{cap},i}
&=\operatorname{smin}_\tau\!\Bigl(v_{\max},\,\sqrt{\tfrac{a_{\text{lat}}^{\text{max}}}{|\kappa_i|+\varepsilon}}\Bigr)
\label{eq:vcap} \\
&= -\tau\log\!\left(
e^{-v_{\max}/\tau}+e^{-\left(\sqrt{\tfrac{a_{\text{lat}}^{\text{max}}}{|\kappa_i|+\varepsilon}}\right)/\tau}
\right).
\notag
\end{align}
where $v_{\max}$ and $a_{\text{lat}}^{\text{max}}$ denote the vehicle's maximum speed and maximum lateral acceleration. 
We use the smooth minimum operator, $\operatorname{smin}$, to ensure differentiability. As $\tau \to 0$, $\operatorname{smin}_\tau$ converges to $\min$.

In parallel, a bumpiness-aware preferred speed for each segment $i$ is obtained from a time–bumpiness trade-off: 
\begin{equation}
v_{\text{pref},i} = \arg\min_{v > 0} \frac{w_{\text{time}}}{v} + w_{\text{bump}}\, b^\alpha_i\, v
\label{eq:vpref_argmin}
\end{equation}
where $w_{\text{time}}, w_{\text{bump}}$ are weights used to balance trade-offs between travel time and predicted vertical acceleration. The first term in the objective encourages shorter travel time, while the second term penalizes high vertical acceleration, taking into account that vertical acceleration scales proportionally to vehicle speed multiplied by the bumpiness estimate. The exponent $\alpha \in \N$ in $b^\alpha_i$ helps rescale the bumpiness prediction. Empirically, setting $\alpha$ to a high value (e.g. $3$ vs. $1$) creates a larger contrast in the speed profile (i.e. slowing down more in bumpy regions and speeding up more in smooth regions).  
Note that the objective is convex in $v$ and the minimizer can be obtained by setting its gradient to $0$, which yields
$
v_{\text{pref},i}=\sqrt{\frac{w_{\text{time}}}{w_{\text{bump}}\,(b^\alpha_i+\varepsilon)}}.
$
As before, to ensure differentiability, we fuse the speed cap and the preferred speed using the smooth minimum operator $v_i = \operatorname{smin}_\tau (v_{\text{cap},i},\,v_{\text{pref},i})$.
%
Let $\bar v_i=\tfrac{1}{2}(v_i+v_{i+1})$ denote the segment speed and $\bar \kappa_i = \tfrac{1}{2}(\kappa_i+\kappa_{i+1})$ denote the segment curvature. Our objective combines a time surrogate, a speed-weighted bump cost, and geometric regularizers:
%
\begin{equation}
\resizebox{\linewidth}{!}{$
\mathcal{J}(\mathcal{C}) =
\underbrace{\sum_{i=1}^{N-1}\frac{\Delta s_i}{\bar v_i}}_{\text{time surrogate}}
+ \lambda_b\underbrace{\sum_{i=1}^{N-1}\bar b_i \bar v_i \Delta s_i}_{\text{speed-weighted bump cost}}
+ \lambda_s\underbrace{\sum_{i=1}^{N-1}(\Delta s_i)^2}_{\text{path smoothness}}
+ \lambda_\kappa\underbrace{\sum_{i=2}^{N-1}\bar\kappa_i^2}_{\text{curvature smoothness}}
$}
\label{eq:trajopt_objective}
\end{equation}
where the weights $\lambda_b, \lambda_s, \lambda_\kappa$ balance the terms.
Note that the objective $\mathcal{J}(\mathcal{C})$ is differentiable with respect to $\mathbf{P}$ thanks to the differentiability of $f_{\texttt{bump}}$ with respect to the query points as well as the differentiability of $\bar{v}_i, \bar \kappa_i, \Delta s_i$ with respect to $\mathbf{P}$. Since the interpolation from $\mathcal{C}$ to $\mathbf{P}$ is differentiable, the objective is differentiable with respect to $\mathcal{C}$, thus enabling gradient-based optimization. 
We optimize only the interior control points using $\operatorname{Adam}$~\cite{kingma2014adam} with gradient clipping, and project them onto the local planning bounds $[x_{\min},x_{\max}]\times[y_{\min},y_{\max}]$. Start and goal (i.e. $c_1, c_M$) are held fixed. 

After the optimization, we perform a 1-D time-scaling along arc length to obtain a dynamically feasible velocity profile following standard trajectory time-parameterization approaches~\cite{bobrow1985time,pham2018new}. 
We enforce
$v\le v_{\max}$, $v^2|\kappa|\le a_{\text{lat}}^{\text{max}}$, and tangential acceleration limits $a\in[-a_{\text{dec}},\,a_{\text{acc}}]$, while retaining the same soft trade-off between time and bump cost to shape the speed profile. 
Finally, we recover yaw from the path tangent and angular rate via $\omega=v\,\kappa$, yielding the outputs $\{\,t,\,\mathbf{P},\,\mathrm{yaw}(t),\,v(t),\,\omega(t)\}$.

\subsubsection{{Tracking Controller}}
We formulate an MPC problem to track the optimized trajectory. Let $x = [p_x, p_y, \theta]^T, u = [v, \omega]^T$, where $p_x, p_y$ denote the $x, y$ coordinates of the vehicle, $\theta$ denotes the yaw angle, and $v, \omega$ denote the linear speed and angular rate.   
The optimization problem is then:
\begin{align*}
    \min_{\substack{x_1,\dots,x_N\\ u_0,\dots,u_{N-1}}}
    \sum_{k=0}^{N-1} & \Big( \,\|\state_k -\xref_k\|_{Q}^2 + \|u_k -\uref_k\|_{R}^2   \\
    &~~+  \|\state_N -\xref_N\|_{Q_N}^2 \Big) \\
   \text{s.t. }~   
  x_{k+1} &
   = x_k 
   + \dt
  \begin{bmatrix}
    \cos(\yaw_k)\,v_k\\
    \sin(\yaw_k)\,v_k \\
    \omega_k
  \end{bmatrix} \\
  v_{\min} & \le v_k \le v_{\max} \\
  \omega_{\min} & \le \omega_k \le \omega_{\max}, ~\forall k \in \{0, \dots, N-1\},
\end{align*}
which is solved using {\bf do-mpc} \cite{fiedler2023mpc,andersson2019casadi,wachter2006implementation}.

\section{Simulation Experiments}
We first evaluate the proposed approach on a series of representative off-road navigation scenarios in a Unity simulation environment.

\subsection{System Baselines}
We evaluate \texttt{TRAIL} against two end-to-end navigation systems that include perception, planning, and control: \texttt{Geometry} and \texttt{WayFASTER}. These comparisons therefore reflect full navigation performance (success rate, execution time, etc.).
For \texttt{Geometry}, we implement a method inspired by STEP \cite{fan2021step} and RoadRunner \cite{frey2024roadrunner}: geometric risk factors (step size, slope, curvature) are fused via Conditional Value at Risk (CVaR) to form a risk map, which is then used by an MPPI controller. Our implementation uses our encoder and the WayFASTER decoder \cite{gasparino2024wayfaster} due to lack of publicly available code. We refer to this baseline as \texttt{Geometry}.
For the traction-based baseline, we use WayFASTER \cite{gasparino2024wayfaster} with its released network architecture\footnote{\url{https://github.com/matval/wayfaster}}, paired with an MPPI controller following the formulation in the original paper.
We implemented both MPPI controllers by adapting a public PyTorch-based codebase\footnote{\url{https://github.com/UM-ARM-Lab/pytorch_mppi}}. We use a horzion $N=20$ with a time step of $dt=0.1\si{\second}$.\\

\begin{figure}[!t]
    \centering
    \includegraphics[trim=0 245 600 0, clip, scale=0.5]{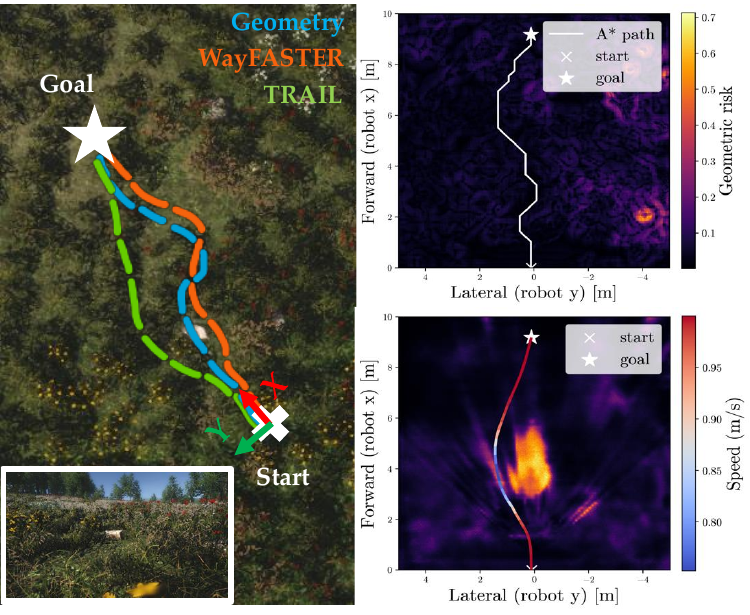}
    \caption{\texttt{Grassland} scenario. The robot’s forward direction ($x$) and leftward direction ($y$) are represented by the red and green axes, respectively. Examples of paths produced by \texttt{Geo-20} (blue), \texttt{WF-20} (orange), and \texttt{TRAIL} (green) are shown on the left. The bottom left corner shows the camera image from the robot. On the right side, the top shows the initial path generated by A$^\star$ overlaid on the geometric risk map. Bottom right shows the optimized trajectory overlaid on the predicted bumpiness value. The trajectory is color-coded by speed.}
    \label{fig:sim_exp_grass_paths}
\end{figure}


\subsection{Dataset and Training}

The training data consists of about 1600$\si{\second}$ of a robot traversing grassland and forest, where it encounters small rocks and bushes intermittently but avoids trees and large obstacles.
As we will see in the results, WayFASTER struggles to learn useful features due to the sparse training signal. 
The optimizer for both our method and \texttt{Geometry} is AdamW~\cite{loshchilov2019decoupled} 
with a learning rate of $5\times10^{-5}$ and a weight decay of $10^{-4}$.
A cosine annealing scheduler~\cite{loshchilov2017sgdr} 
is applied with $T_{\max}=10$ and $\eta_{\min}=10^{-5}$ 
to gradually reduce the learning rate during training.
The optimizer for \texttt{WayFASTER} is as provided in their codebase.
Each method is trained for 100 epochs with a batch size of $4$, and the best checkpoint is used.

 \subsection{Metrics}
To evaluate each method, we repeat every run three times and report the following: Success Rate (percentage of runs reaching the goal), Progress (fraction of initial distance covered), Time (time to reach the goal), Length (total path length), Mean and Max Vertical Acceleration RMS (computed over a $0.1\si{\second}$ window). Runs that exceed the $60\si{\second}$ time cap are marked as failures. All metrics except Success Rate and Progress are computed only on successful runs.



\subsection{Timing Analysis}
Timings are performed on a workstation with an RTX 4090 GPU, an AMD Ryzen 9 7950x 16-core processor with 32 threads, and 128 GB of RAM. The encoder of our model and \texttt{Geometry} take $25 \si{\milli\second}$ per inference and \texttt{WayFASTER}'s encoder takes $20 \si{\milli\second}$. 
With horizon $N = 20$, \texttt{Geometry}'s MPPI takes $10 \si{\milli\second}$ per inference and \texttt{WayFASTER}'s MPPI takes $25 \si{\milli\second}$ per inference. For the proposed method, A$^\star$ takes $5~\si{\milli\second}$ (to generate a path with 40 points), trajectory optimization takes $7.5~\si{\milli\second}$ per iteration (to optimize an initial path with 30 points, which are downsampled from the A$^\star$ path), and the MPC take $6 \si{\milli\second}$ with horizon $N=20$.


\begin{figure}[!t]
    \centering
    \includegraphics[trim=0 68 575 0, clip, scale=0.6]{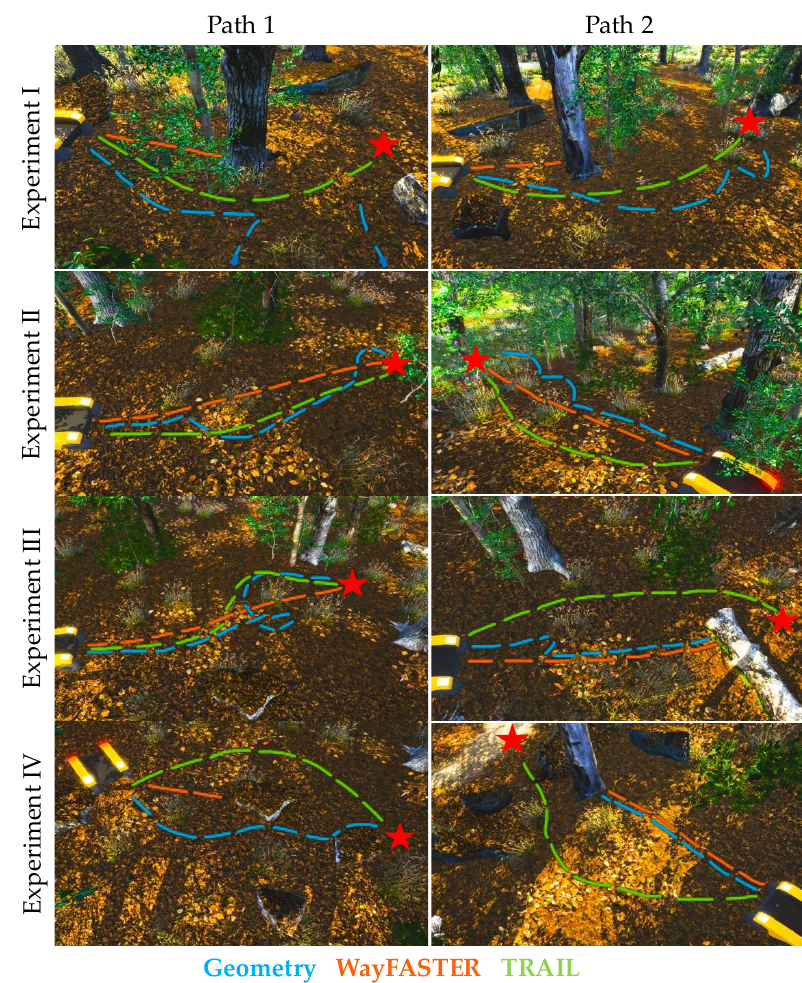}
    \caption{\texttt{Forest} scenario. 
    Each row shows one set of experiments. The first column shows Path 1 and the second column shows Path 2 of each set of experiments. The forest consists of tall grass, bushes, rocks, logs, and trees. Images are tone- and color-corrected for visibility. The proposed method is able to navigate efficiently while the two baselines struggle: \texttt{Geometry} is conservative due to the density of objects with high elevations while \texttt{WayFASTER} is too optimistic since there are no collision events with similar objects in the training data. On the other hand, \texttt{TRAIL} is able to recognize and avoid rocks while not being constrained by the tall grass.
    }
    \label{fig:sim_forest_paths}
\end{figure}

\subsection{Navigation Performance}
\begin{table}[!t]
  \centering
  \scriptsize
  \caption{Comparison Results (\texttt{Grassland})}
  \label{tab:traversability_sim_grass_results}
  \setlength{\tabcolsep}{2pt}
  \renewcommand{\arraystretch}{1.1}
  \resizebox{\columnwidth}{!}{
  \begin{tabular}{lccccc}
  \toprule
  Metric & \texttt{Geo-20} & \texttt{Geo-50} & \texttt{Way-20} & \texttt{Way-50} & \texttt{TRAIL} \\
  \midrule
  Success [\si{\percent}] 
    & \besttt{100.0} & \besttt{100.0} & 66.7 & \besttt{100.0} & \besttt{100.0} \\[2pt]

  Progress [\si{\percent}]  
    & \besttt{100.0 $\pm$ 0.0} & \besttt{100.0 $\pm$ 0.0} & 76.0 $\pm$ 33.9 & \besttt{100.0 $\pm$ 0.0} & \besttt{100.0 $\pm$ 0.0} \\[2pt]

  Time [\si{\second}] 
    & 21.3 $\pm$ 1.9 & 60.5 $\pm$ 0.3 & \second{16.6 $\pm$ 2.3} & 17.2 $\pm$ 2.0 & \besttt{12.0 $\pm$ 0.4} \\[2pt]

  Length [\si{\meter}] 
    & \second{10.7 $\pm$ 0.5} & 14.3 $\pm$ 1.1 & 11.7 $\pm$ 1.2 & 11.9 $\pm$ 0.9 & \besttt{10.3 $\pm$ 0.4} \\[2pt]

  Vert. Acc. RMS [\si{\meter\per\square\second}] 
    & 2.8 $\pm$ 0.1 & 2.7 $\pm$ 0.1 & \second{2.5 $\pm$ 0.3} & 2.6 $\pm$ 0.5 & \besttt{1.8 $\pm$ 0.3} \\[2pt]

  Vert. Acc. Max [\si{\meter\per\square\second}] 
    & 75.8 $\pm$ 6.6 & 72.8 $\pm$ 47.5 & \second{43.4 $\pm$ 23.4} & 58.7 $\pm$ 15.9 & \besttt{4.4 $\pm$ 0.5} \\
  \bottomrule
  \end{tabular}}
\end{table}

\noindent 
{\bf Grassland} 
The \texttt{Grassland} scenario (see \Cref{fig:sim_exp_grass_paths}) features a rock hidden in tall grass (bottom left of \Cref{fig:sim_exp_grass_paths}). The quantitative results in \Cref{tab:traversability_sim_grass_results} show that \texttt{Geometry}, whose inflated geometric map cannot distinguish the rock from tall grass, produces an almost uniform risk map, drives straight toward the goal, runs over the rock, and incurs large vertical acceleration (\texttt{Geo-20}). Because CVaR values lie in $[0,1]$ \cite{frey2024roadrunner}, MPPI’s behavior is highly sensitive to the trade-off between risk and goal distance: high risk weight leads to oscillations and failure to reach the goal, while low risk weight reproduces \texttt{Geometry}’s straight-line behavior, consistent with our ablation study in \Cref{sec:ablation}. \texttt{WayFASTER} detects the rock but is overly optimistic, briefly clipping it and producing high vertical acceleration (\texttt{WF-20}), likely due to weak BEV cues from sparse supervision and a limited encoder. In contrast, our method maintains clearance, using bumpiness during trajectory optimization to push the path away from the rock while remaining less conservative in grass, yielding a smoother, shorter trajectory and an adaptive speed profile that slows near the rock and accelerates through grass. Increasing the MPPI horizon from $20$ to $50$ confirms that the baselines are not horizon-limited: \texttt{Geo-50} becomes indecisive and slower due to accumulated risk, while \texttt{WF-50} matches \texttt{WF-20} (the single \texttt{WF-20} failure resulted from odometry divergence). Thus, we use $N=20$ for subsequent experiments.

\begin{table}[!t]
  \centering
  \tiny
  \renewcommand{\arraystretch}{1.0}
  \setlength{\tabcolsep}{3pt}
  \caption{Comparison Results (\texttt{Forest})}
  \label{tab:traversability_sim_forest_results_compact}
  \begin{tabular}{c l c c c c c c}
    \toprule
      & Method
      & Success (\%)
      & Progress (\%)
      & Time (s)
      & Length (m)
      & \multicolumn{2}{c}{Acc (m/s$^2$)} \\
      &
      &
      &
      &
      &
      &
      RMS
      & Max \\
    \midrule

    \multicolumn{8}{c}{\textbf{Forest Experiment I}} \\
    \midrule

    \multirow{3}{*}{\rotatebox[origin=c]{90}{{Path 1}}}
      & \texttt{Geometry}
      & \besttt{100.0}
      & \besttt{100.0 $\pm$ 0.0}
      & \second{44.0 $\pm$ 1.6}
      & \second{24.0 $\pm$ 1.5}
      & \second{1.8 $\pm$ 0.1}
      & \second{6.1 $\pm$ 0.3} \\
      & \texttt{WayFASTER}
      & 0.0
      & 60.8 $\pm$ 4.6
      & N/A & N/A & N/A & N/A \\
      & \texttt{TRAIL}
      & \besttt{100.0}
      & \besttt{100.0 $\pm$ 0.0}
      & \besttt{11.5 $\pm$ 2.2}
      & \besttt{11.4 $\pm$ 2.2}
      & \besttt{1.7 $\pm$ 0.1}
      & \besttt{6.0 $\pm$ 1.7} \\
    \addlinespace[0.4ex]

    \multirow{3}{*}{\rotatebox[origin=c]{90}{{Path 2}}}
      & \texttt{Geometry}
      & \besttt{100.0}
      & \besttt{100.0 $\pm$ 0.0}
      & \second{38.3 $\pm$ 14.2}
      & \second{20.9 $\pm$ 9.0}
      & \second{1.8 $\pm$ 0.1}
      & \second{6.2 $\pm$ 0.1} \\
      & \texttt{WayFASTER}
      & 0.0
      & 31.3 $\pm$ 2.3
      & N/A & N/A & N/A & N/A \\
      & \texttt{TRAIL}
      & \besttt{100.0}
      & \besttt{100.0 $\pm$ 0.0}
      & \besttt{10.9 $\pm$ 1.1}
      & \besttt{10.8 $\pm$ 1.0}
      & \besttt{1.7 $\pm$ 0.1}
      & \besttt{5.1 $\pm$ 0.7} \\
    \midrule

    \multicolumn{8}{c}{\textbf{Forest Experiment II}} \\
    \midrule

    \multirow{3}{*}{\rotatebox[origin=c]{90}{{Path 1}}}
      & \texttt{Geometry}
      & \besttt{100.0}
      & \besttt{100.0 $\pm$ 0.0}
      & 26.8 $\pm$ 6.7
      & 14.2 $\pm$ 1.3
      & 1.9 $\pm$ 0.2
      & \besttt{5.5 $\pm$ 0.5} \\
      & \texttt{WayFASTER}
      & \besttt{100.0}
      & \besttt{100.0 $\pm$ 0.0}
      & \second{11.5 $\pm$ 1.8}
      & \second{10.9 $\pm$ 0.9}
      & \second{1.8 $\pm$ 0.2}
      & \second{5.5 $\pm$ 0.8} \\
      & \texttt{TRAIL}
      & \besttt{100.0}
      & \besttt{100.0 $\pm$ 0.0}
      & \besttt{10.9 $\pm$ 0.4}
      & \besttt{10.9 $\pm$ 0.4}
      & \besttt{1.8 $\pm$ 0.1}
      & 6.0 $\pm$ 0.3 \\
    \addlinespace[0.4ex]

    \multirow{3}{*}{\rotatebox[origin=c]{90}{{Path 2}}}
      & \texttt{Geometry}
      & 33.3
      & 91.3 $\pm$ 6.2
      & 45.3 $\pm$ 0.0
      & 22.3 $\pm$ 0.0
      & 1.9 $\pm$ 0.0
      & 6.2 $\pm$ 0.0 \\
      & \texttt{WayFASTER}
      & \besttt{100.0}
      & \besttt{100.0 $\pm$ 0.0}
      & \second{13.3 $\pm$ 1.7}
      & \besttt{12.1 $\pm$ 0.8}
      & \second{1.9 $\pm$ 0.1}
      & \second{5.8 $\pm$ 0.2} \\
      & \texttt{TRAIL}
      & \besttt{100.0}
      & \besttt{100.0 $\pm$ 0.0}
      & \besttt{12.3 $\pm$ 0.7}
      & \second{12.5 $\pm$ 1.0}
      & \besttt{1.7 $\pm$ 0.1}
      & \besttt{5.5 $\pm$ 0.5} \\
    \midrule

    \multicolumn{8}{c}{\textbf{Forest Experiment III}} \\
    \midrule

    \multirow{3}{*}{\rotatebox[origin=c]{90}{{Path 1}}}
      & \texttt{Geometry}
      & \besttt{100.0}
      & \besttt{100.0 $\pm$ 0.0}
      & 52.9 $\pm$ 10.4
      & 19.1 $\pm$ 3.0
      & \besttt{1.7 $\pm$ 0.1}
      & 14.3 $\pm$ 12.7 \\
      & \texttt{WayFASTER}
      & \besttt{100.0}
      & \besttt{100.0 $\pm$ 0.0}
      & \besttt{10.8 $\pm$ 0.1}
      & \besttt{10.4 $\pm$ 0.1}
      & 1.8 $\pm$ 0.0
      & \second{5.7 $\pm$ 0.4} \\
      & \texttt{TRAIL}
      & \besttt{100.0}
      & \besttt{100.0 $\pm$ 0.0}
      & \second{10.9 $\pm$ 0.3}
      & \second{11.0 $\pm$ 0.2}
      & \besttt{1.7 $\pm$ 0.1}
      & \besttt{4.5 $\pm$ 0.7} \\
    \addlinespace[0.4ex]

    \multirow{3}{*}{\rotatebox[origin=c]{90}{{Path 2}}}
      & \texttt{Geometry}
      & 0.0
      & \second{67.0 $\pm$ 1.2}
      & N/A & N/A & N/A & N/A \\
      & \texttt{WayFASTER}
      & 0.0
      & 61.3 $\pm$ 1.5
      & N/A & N/A & N/A & N/A \\
      & \texttt{TRAIL}
      & \besttt{66.7}
      & \besttt{91.5 $\pm$ 12.1}
      & \besttt{12.9 $\pm$ 0.5}
      & \besttt{11.6 $\pm$ 0.1}
      & \besttt{2.0 $\pm$ 0.4}
      & \besttt{31.6 $\pm$ 37.0} \\
    \midrule

    \multicolumn{8}{c}{\textbf{Forest Experiment IV}} \\
    \midrule

    \multirow{3}{*}{\rotatebox[origin=c]{90}{{Path 1}}}
      & \texttt{Geometry}
      & \second{66.7}
      & \second{79.1 $\pm$ 29.5}
      & \second{33.5 $\pm$ 11.8}
      & \second{13.7 $\pm$ 1.3}
      & \second{4.0 $\pm$ 1.7}
      & \second{79.3 $\pm$ 54.0} \\
      & \texttt{WayFASTER}
      & 0.0
      & 14.0 $\pm$ 14.0
      & N/A & N/A & N/A & N/A \\
      & \texttt{TRAIL}
      & \besttt{100.0}
      & \besttt{100.0 $\pm$ 0.0}
      & \besttt{12.7 $\pm$ 0.2}
      & \besttt{12.5 $\pm$ 0.2}
      & \besttt{1.9 $\pm$ 0.2}
      & \besttt{5.1 $\pm$ 0.8} \\
    \addlinespace[0.4ex]

    \multirow{3}{*}{\rotatebox[origin=c]{90}{{Path 2}}}
      & \texttt{Geometry}
      & 0.0
      & \second{60.4 $\pm$ 4.2}
      & N/A & N/A & N/A & N/A \\
      & \texttt{WayFASTER}
      & 0.0
      & 57.6 $\pm$ 1.5
      & N/A & N/A & N/A & N/A \\
      & \texttt{TRAIL}
      & \besttt{100.0}
      & \besttt{100.0 $\pm$ 0.0}
      & \besttt{15.3 $\pm$ 1.9}
      & \besttt{14.7 $\pm$ 1.3}
      & \besttt{3.6 $\pm$ 0.3}
      & \besttt{52.2 $\pm$ 17.8} \\
    \bottomrule
  \end{tabular}
\end{table}

\noindent 
{\bf Forest}
To further investigate the performance of the different methods, we pick 8 paths in a forest that represent different challenges, grouped into 4 sets of experiments, as shown in \Cref{fig:sim_forest_paths}. The forest has trees, rocks, fallen logs, bushes, and tall grass. However, unlike in the training dataset, most rocks in the forest are covered with leaves, making them visually indistinguishable from the ground. 
Across the four sets of experiments, the baseline methods exhibit distinct failure modes: in Experiment I, \texttt{Geometry} reaches the goal but takes long, oscillatory detours around tall grass, whereas \texttt{WayFASTER} fails entirely by classifying the tree as traversable due to lack of similar training samples. In Experiment II, all methods reach the goal, but \texttt{Geometry} again oscillates in tall grass, while our method takes a slightly longer path yet achieves higher average speed and lower overall vertical acceleration (recall that vertical acceleration scales proportionally to vehicle speed). In Experiment III, \texttt{WayFASTER} incorrectly treats the log as traversable, and \texttt{Geometry} incorrectly treats the surrounding grass and bush as untraversable—leaving no viable path—whereas our method succeeds in 2 of 3 trials, with the single failure caused by an odometry divergence after a wheel struck the tree trunk. In Experiment IV, \texttt{WayFASTER} fails to detect leaf-covered rocks and drives into the ``bug trap'', while \texttt{Geometry} becomes overly conservative and also gets stuck. Our method avoids the trap (Path 1), identifies the small clearing around the tree (Path 2), and reaches the goal, with vertical acceleration mainly attributable to unavoidable low rocks.

\subsection{Ablation Study}\label{sec:ablation}
\begin{table}[!th]
  \centering
  \scriptsize
  \caption{Ablation Study Results (\texttt{Grassland})}
  \label{tab:traversability_ablation_sim_grass_results}
  \setlength{\tabcolsep}{2pt}
  \renewcommand{\arraystretch}{1.1}
  \resizebox{\columnwidth}{!}{
  \begin{tabular}{lcccccc}
  \toprule
  Metric &
  \rotatebox{45}{\texttt{TRAIL}} &
  \rotatebox{45}{\texttt{MPPI-Geo}} &
  \rotatebox{45}{\texttt{MPPI-Geo-Term}} &
  \rotatebox{45}{\texttt{MPPI-Bump}} &
  \rotatebox{45}{\texttt{MPPI-A$^\star$-Bump}} &
  \rotatebox{45}{\texttt{MPPI-Fused}} \\
  \midrule
  Success [\si{\percent}]                    & \besttt{100.0}   & 0.0   & \besttt{100.0} & \besttt{100.0} & \besttt{100.0} & \besttt{100.0} \\[2pt]
  Progress [\si{\percent}]                   & \besttt{100.0 $\pm$ 0.0} & 9.1 $\pm$ 3.2   & \besttt{100.0 $\pm$ 0.0} & \besttt{100.0 $\pm$ 0.0} & \besttt{100.0 $\pm$ 0.0} & \besttt{100.0 $\pm$ 0.0} \\[2pt]
  Time [\si{\second}]                        & \besttt{12.0 $\pm$ 0.4}   & N/A    & {17.3 $\pm$ 2.2} & \second{17.2 $\pm$ 0.4} & 54.8 $\pm$ 4.6 & 35.6 $\pm$ 18.4 \\[2pt]
  Length [\si{\meter}]                       & \besttt{10.3 $\pm$ 0.4}   & N/A    & 11.1 $\pm$ 0.7 & \second{10.6 $\pm$ 0.2} & 15.5 $\pm$ 0.6 & 14.2 $\pm$ 3.5 \\[2pt]
  Vert. Acc. RMS [\si{\meter\per\square\second}] & \besttt{1.8 $\pm$ 0.3} & N/A    & 3.0 $\pm$ 0.3 & \second{2.0 $\pm$ 0.3} & 1.9 $\pm$ 0.2 & 2.8 $\pm$ 0.6 \\[2pt]
  Vert. Acc. Max [\si{\meter\per\square\second}] & \besttt{4.4 $\pm$ 0.5} & N/A    & 61.4 $\pm$ 26.6 & \second{21.5 $\pm$ 23.7} & 41.6 $\pm$ 26.1 & 87.3 $\pm$ 9.2 \\
  \bottomrule
  \end{tabular}}
\end{table}
To further evaluate the proposed planner, we conduct an ablation study of different MPPI formualtions. 
We generate fixed-resolution cost maps from our trained model and using MPPI as the planner, following common practice. We implement five variants:
\begin{itemize}
\item \texttt{MPPI-Geo}: uses the same geometric cost map as A$^\star$.
\item \texttt{MPPI-Geo-Term}: same as \texttt{MPPI-Geo}, but with a $5 \times$ higher terminal-cost weight.
\item \texttt{MPPI-Bump}: uses the predicted bumpiness as the cost.
\item \texttt{MPPI-A$^\star$-Bump}: initializes MPPI with an A$^\star$ path while minimizing bump cost.
\item \texttt{MPPI-Fused}: averages the geometric and bumpiness maps to form a fused cost.
\end{itemize}
In summary, each MPPI formulation minimizes the distance to goal and cost along the path, with a small penalty on control effort. \texttt{MPPI-A$^\star$-Bump} has an additional cost term that penalizes deviations from the initial path. The planning horizon is set to $N=20$ with $dt = 0.1s$ for all variants.
Note that the smoothness terms in \Cref{eq:trajopt_objective} are not necessary for MPPI since dynamic feasibility is enforced by sampling bounded control inputs and rolling out the dynamics model.

The result is recorded in \Cref{tab:traversability_ablation_sim_grass_results}. \texttt{MPPI-Geo} is unable to reach the goal due to the high geometric risk. On the other hand, \texttt{MPPI-Geo-Term} is able to reach the goal due to a higher terminal-cost weight. However, it is unable to distinguish between traversable vs. un-traversable areas and ends up going straight towards the goal, experiencing large vertical acceleration. Note that the behavior of \texttt{MPPI-Geo-Term} is similar to \texttt{Geo-20} (see \Cref{tab:traversability_sim_grass_results}). 
\texttt{MPPI-Bump} performs the second best, exhibiting similar behavior to \texttt{Way-20}. However, it has a much lower max vertical acceleration, due to the better prediction from our network. 
\texttt{MPPI-Geo-Term} takes significantly longer to reach the goal due to its conflicting objectives. Similarly, \texttt{MPPI-Fused} takes more time to reach the goal while experiencing large vertical acceleration. \\

\begin{table}[!t]
  \centering
  \tiny
  \renewcommand{\arraystretch}{1.0}
  \setlength{\tabcolsep}{3pt}
  \caption{Additional Simulation Investigation Results}
  \label{tab:extra_sim_benchmark}
  {
  \begin{tabular}{c l c c c c c c}
    \toprule
      & Method
      & Success (\%)
      & Progress (\%)
      & Time (s)
      & Length (m)
      & \multicolumn{2}{c}{Acc (m/s$^2$)} \\
      &
      &
      &
      &
      &
      &
      RMS
      & Max \\
    \midrule

    \multicolumn{8}{c}{\textbf{Grassland}} \\
    \midrule

    & \texttt{MPPI-A*-Bump} & 
    100.0  &
    100.0 $\pm$ 0.0 &
    54.8 $\pm$ 4.6 &
    15.5 $\pm$ 0.6 &
    1.9 $\pm$ 0.2 &
    41.6 $\pm$ 26.1 \\
    & \texttt{SALON}        & 
    100.0 &
    100.0 $\pm$ 0.0 &
    27.8 $\pm$ 4.6 &
    13.0 $\pm$ 2.6 &
    2.1 $\pm$ 0.1 &
    60.4 $\pm$ 42.3\\
    & \texttt{SDF}      & 
    100.0 &
    100.0 $\pm$ 0.0 &
    17.7 $\pm$ 1.4 &
    13.9 $\pm$ 0.7 &
    3.4 $\pm$ 1.1 &
    53.4 $\pm$ 34.4 \\
    \midrule

    \multicolumn{8}{c}{\textbf{Forest Experiment I}} \\
    \midrule

    \multirow{3}{*}{\rotatebox[origin=c]{90}{{Path 1}}}
    & \texttt{MPPI-A*-Bump} & 
                   66.7 &
                   91.9 $\pm$ 11.5 &
                   19.7 $\pm$ 2.4 &
                   12.3 $\pm$ 1.3 &
                   2.3 $\pm$ 0.3 &
                   9.6 $\pm$ 4.1\\
                & \texttt{SALON} &
                    0.0 &
                    64.4 $\pm$ 1.8 &
                    N/A &
                    N/A &
                    N/A &
                    N/A \\
                & \texttt{SDF} &
                100.0 &
                100.0 $\pm$ 0.0 &
                16.1 $\pm$ 4.1 &
                14.6 $\pm$ 3.0 &
                1.6 $\pm$ 0.1 &
                5.3 $\pm$ 0.5 \\
    \addlinespace[0.6ex]
    \multirow{3}{*}{\rotatebox[origin=c]{90}{{Path 2}}} 
    & \texttt{MPPI-A*-Bump} &
                    66.7 &
                    76.6 $\pm$ 33.0 &
                    35.5 $\pm$ 17.2 &
                    14.9 $\pm$ 4.0 &
                    1.8 $\pm$ 0.4 &
                    5.6 $\pm$ 0.7 \\
                & \texttt{SALON} &
                    33.3 &
                    51.5 $\pm$ 34.7 &
                    17.00 $\pm$ 0.0 &
                    10.4 $\pm$ 0.0 &
                    1.9 $\pm$ 0.5 &
                    4.2 $\pm$ 0.5 \\
                & \texttt{SDF} &
                66.7 &
                66.7 $\pm$ 47.1 &
                18.4 $\pm$ 0.3 &
                17.0 $\pm$ 0.4 &
                2.2 $\pm$ 0.4 &
                5.4 $\pm$ 0.7 \\
    \addlinespace[0.6ex]

    \midrule

    \multicolumn{8}{c}{\textbf{Forest Experiment II}} \\
    \midrule

    \multirow{3}{*}{\rotatebox[origin=c]{90}{{Path 1}}}
    & \texttt{MPPI-A*-Bump} & 
                   66.7 &
                   66.7 $\pm$ 47.1 &
                   36.5 $\pm$ 5.4 &
                   14.8 $\pm$ 1.7 &
                   3.2 $\pm$ 2.0 &
                   53.8 $\pm$ 66.0 \\
                & \texttt{SALON} &
                    0.0 &
                    20.0 $\pm$ 1.6 &
                    N/A &
                    N/A &
                    N/A &
                    N/A \\
                & \texttt{SDF} &
                66.7 &
                98.0 $\pm$ 2.8 &
                12.6 $\pm$ 0.7 &
                10.7 $\pm$ 0.3 &
                1.9 $\pm$ 0.1 &
                6.8 $\pm$ 1.0 \\
    \addlinespace[0.6ex]
    \multirow{3}{*}{\rotatebox[origin=c]{90}{{Path 2}}}
    & \texttt{MPPI-A*-Bump} &
                    100.0 & 
                    100.0 $\pm$ 0.0 &
                    22.7 $\pm$ 4.2 &
                    13.0 $\pm$ 2.0 &
                    1.9 $\pm$ 0.2 &
                    5.5 $\pm$ 0.4 \\
                & \texttt{SALON} &
                    100.0 &
                    100.0 $\pm$ 0.0 &
                    28.2 $\pm$ 3.1 &
                    13.6 $\pm$ 0.8 &
                    1.7 $\pm$ 0.0 &
                    5.4 $\pm$ 0.5\\
                & \texttt{SDF} &
                66.7 &
                97.6 $\pm$ 3.4 &
                15.9 $\pm$ 0.9 &
                14.0 $\pm$ 1.6 &
                1.8 $\pm$ 0.2 &
                5.4 $\pm$ 0.6 \\
    \addlinespace[0.6ex]
    \midrule

    \multicolumn{8}{c}{\textbf{Forest Experiment III}} \\
    \midrule

    \multirow{3}{*}{\rotatebox[origin=c]{90}{{Path 1}}}
    & \texttt{MPPI-A*-Bump} & 
                   33.3 &
                   91.8 $\pm$ 5.8 &
                   16.2 $\pm$ 0.0 &
                   10.8 $\pm$ 0.0 &
                   2.0 $\pm$ 0.3 &
                   5.1 $\pm$ 0.6 \\
                & \texttt{SALON} &
                    0.0 &
                    55.6 $\pm$ 23.7 &
                    N/A &
                    N/A &
                    N/A &
                    N/A \\
                & \texttt{SDF} &
                33.3 &
                89.9 $\pm$ 7.3 &
                16.7 $\pm$ 0.0 &
                14.4 $\pm$ 0.0 &
                2.0 $\pm$ 0.1 &
                6.2 $\pm$ 0.4 \\
    \addlinespace[0.6ex]
    \multirow{3}{*}{\rotatebox[origin=c]{90}{{Path 2}}}
    & \texttt{MPPI-A*-Bump} &
                    33.3 &
                    58.6 $\pm$ 42.6 &
                    17.6 $\pm$ 0.0 &
                    11.7 $\pm$ 0.0 &
                    4.1 $\pm$ 2.4 &
                    129.2 $\pm$ 102.9\\
                & \texttt{SALON} &
                    0.0 &
                    76.1 $\pm$ 3.4 &
                    N/A &
                    N/A &
                    N/A &
                    N/A \\
                & \texttt{SDF} &
                0.0 &
                50.2 $\pm$ 31.4 &
                N/A &
                N/A &
                N/A &
                N/A \\
    \addlinespace[0.6ex]
    \midrule

    \multicolumn{8}{c}{\textbf{Forest Experiment IV}} \\
    \midrule

    \multirow{3}{*}{\rotatebox[origin=c]{90}{{Path 1}}}
    & \texttt{MPPI-A*-Bump} & 
                  33.3 &
                  61.6 $\pm$ 32.9 &
                  24.2 $\pm$ 0.0 &
                  12.6 $\pm$ 0.0 &
                  2.4 $\pm$ 0.7 &
                  54.4 $\pm$ 36.5 \\
                & \texttt{SALON} &
                    0.0 &
                    74.6 $\pm$ 2.3 &
                    N/A &
                    N/A &
                    N/A &
                    N/A \\
                & \texttt{SDF} &
                66.7 &
                97.7 $\pm$ 3.2 &
                14.8 $\pm$ 0.4 &
                12.5 $\pm$ 0.5 &
                2.2 $\pm$ 0.1 &
                19.0 $\pm$ 2.9 \\
    \addlinespace[0.6ex]
    \multirow{3}{*}{\rotatebox[origin=c]{90}{{Path 2}}}
    & \texttt{MPPI-A*-Bump} &
                   0.0 &
                   76.5 $\pm$ 11.9 &
                   N/A &
                   N/A &
                   N/A &
                   N/A \\
                & \texttt{SALON} &
                    0.0 &
                    45.3 $\pm$ 36.6 &
                    N/A &
                    N/A &
                    N/A &
                    N/A\\
                & \texttt{SDF} &
                0.0 &
                55.6 $\pm$ 10.6 &
                N/A &
                N/A &
                N/A &
                N/A \\
    \addlinespace[0.6ex]
    \bottomrule
  \end{tabular}
  }
\end{table}
\subsection{Additional Simulation Investigations}\label{sec:ablation_planner}
In addition to the end-to-end system baselines (\texttt{Geometry}, \texttt{WayFASTER}), we report simulation-only investigations to isolate specific factors: (i) the effect of global A$^\star$ initialization (\texttt{MPPI-A$^\star$-Bump}), (ii) a trajectory-refinement variant that optimizes obstacle clearance/smoothness without bumpiness-driven speed adaptation (\texttt{SDF} with the same A$^\star$ initialization as \texttt{TRAIL}), and (iii) a small-data traversability learning comparison (\texttt{SALON} \cite{sivaprakasam2025salon}), where we evaluate the released perception module paired with a custom MPPI controller due to the absence of a released planner formulation. The custom MPPI minimizes rollout cost from {SALON}'s cost map while enforcing the speed constraint derived from {SALON}'s speed map, consistent with the high-level description in \cite{sivaprakasam2025salon}. Results are reported in \cref{tab:extra_sim_benchmark}.

Overall, \texttt{MPPI-A$^\star$-Bump} is sensitive to competing objective terms and exhibits oscillatory/indecisive behavior in our settings, leading to longer execution times (\Cref{tab:extra_sim_benchmark}). 
\texttt{SDF} performs reasonably well in scenarios dominated by geometric obstacle avoidance, but degrades in cases involving tall vegetation, where obstacle geometry alone is ambiguous.
\texttt{SALON} improves over \texttt{WayFASTER} in several small-data cases, but can miss hazards that are underrepresented in (or visually different from) its experience buffer, leading to failures in our forest scenarios. 
For example, \texttt{SALON} successfully assigns high cost to the rock region in \texttt{Grassland}, but can fail to detect leaf-covered rocks or trees in darker forest regions (e.g. Experiment IV Path 1 in \Cref{fig:sim_forest_paths}).

\section{Field Experiments}
In hardware experiments, we focus on end-to-end system comparisons (\texttt{Geometry} and \texttt{WayFASTER}); simulation-only diagnostic studies (e.g., \texttt{SALON}, \texttt{SDF}, \texttt{MPPI-A$^\star$-Bump}) are reported in Sec.~\ref{sec:ablation_planner}.
Our test platform is an AgileX Scout Mini equipped with a ZED~2i stereo camera and an Ouster OS1-32-GEN2.0 LiDAR. The training set contains fewer than 2000 samples collected in a similar area using a Clearpath Husky A300 UGV with the same sensor stack and calibration. The Husky data covers dirt, paved road, tall grass, and uneven terrain with roots and branches, without purposeful collisions with trees.


{
During data collection, Scout Mini odometry drifted substantially (despite post-processing), making label generation unreliable. The Husky’s smoother motion and traction yielded more stable odometry for offline aggregation and annotation. Training on Husky data and deploying on Scout Mini thus introduces a platform distribution shift (e.g., mass and vibration scale), providing an additional generalizability test. In practice, this shift did not noticeably degrade \texttt{TRAIL}: geometric outputs are expressed in the LiDAR frame, and thus transfer naturally across platforms when the sensor stack and calibration are unchanged, and planning relies mainly on relative variations in predicted bumpiness to shape the speed profile. We use KISS-ICP \cite{vizzo2023kiss} for offline odometry in data processing and DLIO \cite{chen2022direct} for higher-rate odometry during deployment.
}

\begin{figure}[!t]
    \centering
    \includegraphics[trim=0 275 00 0, clip, width=0.49\textwidth]{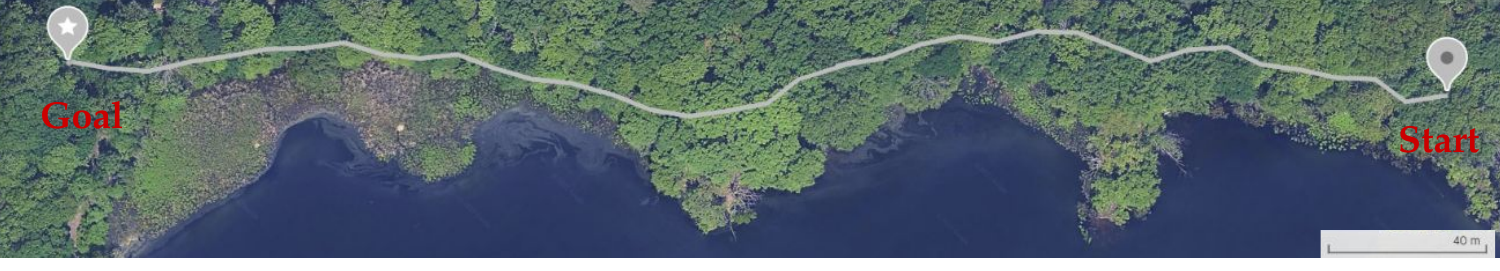}
    \caption{GPS locations of the robot during the long path navigation hardware experiment. The total length is about $370\si{\meter}$.}
    \label{fig:hardware_exp_long_path}
\end{figure}

\begin{table}[t]
  \centering
  \tiny
  \renewcommand{\arraystretch}{1.0}
  \setlength{\tabcolsep}{4pt}
  \caption{Hardware Short Path Navigation Comparison Results 
  the Euclidean distances between start and goal are about \SI{9}{m})}
  \label{tab:traversability_hardware_benchmark_results}

  \begin{tabular}{
    l
    c  
    c  
    c  
    c  
    p{0.85cm}  
    p{0.85cm}  
  }
  \toprule
  \multirow{2}{*}{Method}
    & {Success (\%)} 
    & {Progress (\%)} 
    & {Time (\si{\second})} 
    & {Length (\si{\meter})} 
    & {\begin{tabular}{c}
        Vert Acc \\
        RMS \\
        (\si{\meter\per\square\second})
        \end{tabular}}
    & {\begin{tabular}{c}
        Vert Acc \\
        Max \\
        (\si{\meter\per\square\second})
        \end{tabular}} \\
  \midrule

    \multicolumn{7}{c}{\textbf{Experiment I}} \\
    \midrule
    \texttt{Geometry} 
      & \second{33.3} 
      & \second{75.7 $\pm$ 17.2}
      & \second{28.7 $\pm$ 0.0}
      & \besttt{12.7 $\pm$ 0.0}
      & \besttt{2.0 $\pm$ 0.0}
      & \second{18.4 $\pm$ 0.0} \\
    \texttt{WayFASTER}
      & 0.0
      & 72.0 $\pm$ 10.3
      & N/A
      & N/A
      & ~\quad N/A
      & ~\quad N/A \\
    \texttt{TRAIL}
      & \besttt{100.0}
      & \besttt{100.0 $\pm$ 0.0}
      & \besttt{16.6 $\pm$ 1.0}
      & \second{13.6 $\pm$ 0.9}
      & \second{2.8 $\pm$ 0.2}
      & \besttt{11.9 $\pm$ 1.5} \\
    \midrule

    \multicolumn{7}{c}{\textbf{Experiment II}} \\
    \midrule
    \texttt{Geometry} 
      & 0.0
      & 60.0 $\pm$ 9.3
      & N/A
      & N/A
      & ~\quad N/A
      & ~\quad N/A \\
    \texttt{WayFASTER}
      & \second{33.3}
      & \second{72.2 $\pm$ 19.7}
      & \second{19.3 $\pm$ 0.0}
      & \second{15.4 $\pm$ 0.0}
      & \besttt{1.9 $\pm$ 0.0}
      & \besttt{7.1 $\pm$ 0.0} \\
    \texttt{TRAIL}
      & \besttt{66.7}
      & \besttt{85.8 $\pm$ 20.1}
      & \besttt{16.2 $\pm$ 0.0}
      & \besttt{15.3 $\pm$ 0.6}
      & \second{1.9 $\pm$ 0.1}
      & \second{8.9 $\pm$ 0.6} \\
    \midrule

\multicolumn{7}{c}{\textbf{Experiment III}} \\
\midrule
\texttt{Geometry} 
  & 0.0
  & 40.1 $\pm$ 3.1
  & N/A
  & N/A
  & ~\quad N/A
  & ~\quad N/A \\
\texttt{WayFASTER}
  & 0.0
  & 37.4 $\pm$ 0.4
  & N/A
  & N/A
  & ~\quad N/A
  & ~\quad N/A \\
\texttt{TRAIL}
  & \besttt{66.7}
  & \besttt{91.1 $\pm$ 12.5}
  & \besttt{14.7 $\pm$ 0.6}
  & \besttt{11.1 $\pm$ 0.9}
  & \besttt{2.1 $\pm$ 0.2}
  & \besttt{11.9 $\pm$ 2.6} \\
\midrule

\multicolumn{7}{c}{\textbf{Experiment IV}} \\
\midrule
\texttt{Geometry} 
  & \besttt{100.0}
  & \besttt{100.0 $\pm$ 0.0}
  & \second{28.2 $\pm$ 5.1}
  & \second{13.4 $\pm$ 1.4}
  & \besttt{2.0 $\pm$ 0.3}
  & \second{11.0 $\pm$ 1.5} \\
\texttt{WayFASTER}
  & 0.0
  & 43.0 $\pm$ 19.9
  & N/A
  & N/A
  & ~\quad N/A
  & ~\quad N/A \\
\texttt{TRAIL}
  & \besttt{100.0}
  & \besttt{100.0 $\pm$ 0.0}
  & \besttt{16.3 $\pm$ 1.2}
  & \besttt{11.5 $\pm$ 0.2}
  & \second{2.9 $\pm$ 0.3}
  & \besttt{10.8 $\pm$ 1.5} \\
\midrule
  
\multicolumn{7}{c}{\textbf{Experiment V}} \\
\midrule
\texttt{Geometry} 
  & 0.0
  & 14.5 $\pm$ 14.5
  & N/A
  & N/A
  & ~\quad N/A
  & ~\quad N/A \\
\texttt{WayFASTER}
  & 0.0
  & \second{67.6 $\pm$ 23.5}
  & N/A
  & N/A
  & ~\quad N/A
  & ~\quad N/A \\
\texttt{TRAIL}
  & \besttt{100.0}
  & \besttt{100.0 $\pm$ 0.0}
  & \besttt{24.2 $\pm$ 1.7}
  & \besttt{18.6 $\pm$ 1.1}
  & \besttt{1.4 $\pm$ 0.0}
  & \besttt{7.1 $\pm$ 0.2} \\
  \bottomrule
  \end{tabular}
\end{table}

\begin{figure*}[t]
    \centering
    \includegraphics[trim=0 195 0 5, clip, width=0.98\textwidth]{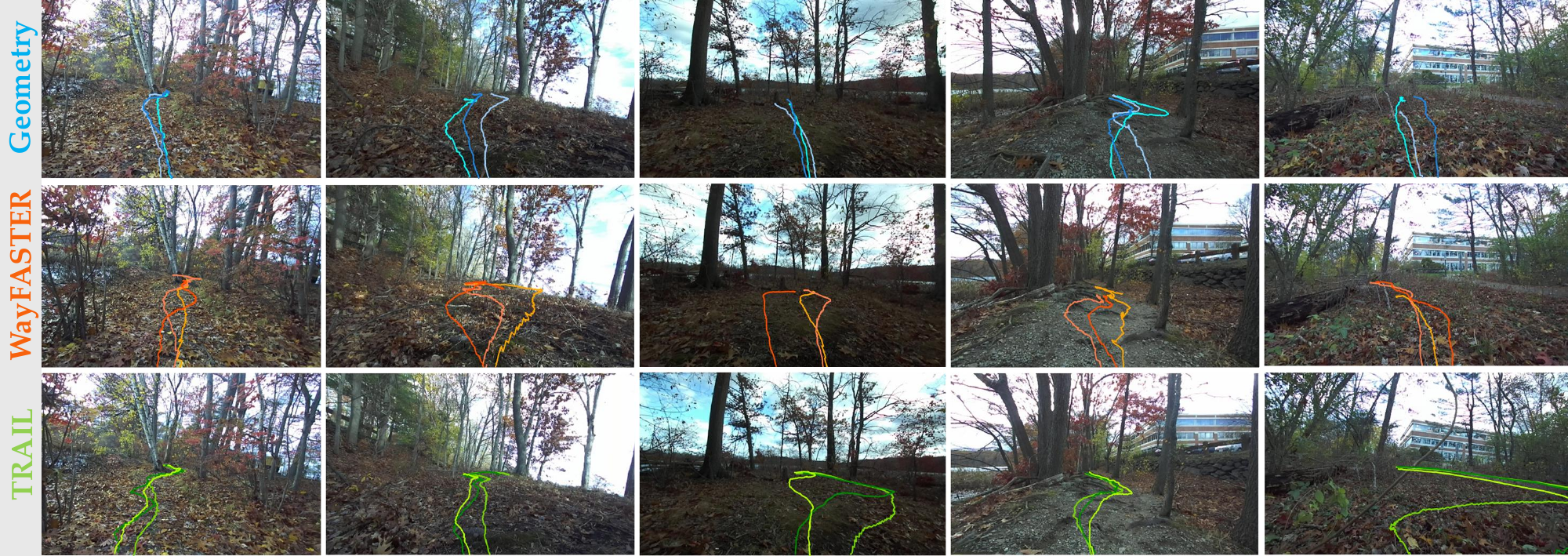}
    \caption{\textbf{Short Path Navigation Results}. Each column shows one set of experiments with separate trials colored differently. The first row shows the paths followed by \texttt{Geometry}, the second row shows the paths followed by \texttt{WayFASTER}, and the third row shows the paths followed by \texttt{TRAIL}.
    The visualizations are obtained by projecting the odometry to the onboard camera image captured at the starting position. In the fifth experiment (last column), \texttt{TRAIL} was able to find a clearing on the right (paved road) and successfully navigate to the goal, while the two baseline methods were trapped by twigs (which are hard to distinguish from the dry grass).}
    \label{fig:short_navigation_visualization}
\end{figure*}

\subsection{Short Path Navigation}

We chose five sets of experiments to test the capability of all methods in scenarios that present different challenges. Each set of experiments consists of three trials per method. A trial is considered failed if the vehicle ends up in an unrecoverable configuration (e.g. in collision with a tree trunk), in which case we manually terminated the trial. 
We set the planning horizon of both baseline approaches to $5.0\si{\second}$ (used in \cite{frey2024roadrunner} for larger scale experiments) and the number of MPPI samples to 4096. 
Quantitative results are recorded in \Cref{tab:traversability_hardware_benchmark_results} and visualizations are shown in \Cref{fig:short_navigation_visualization}. 
Experiments I and IV are challenging due to tree trunks and roots, Experiment II involves tall vegetation, and Experiments III and V combine both scenarios.

\texttt{Geometry} failed all experiments with tall but traversable vegetation, exhibiting behavior similar to that observed in the simulation study. It succeeded in only one of three trials in Experiment I, primarily due to indecisive behavior, oscillating between left- and right-side avoidance (from MPPI sampling trajectories on both sides of the tree).
\texttt{Geometry} was able to navigate successfully in Experiment IV, but it had a longer travel time due to its indecisive behavior near the goal, which was located in a region with high geometric risk. 

\texttt{WayFASTER} failed all the experiments with geometric obstacles, which is consistent with the simulation study. It only succeeded in one of three trials in Experiment II (colored with yellow), where the other trials failed due to collision with a non-traversable thin tree. 

We observe that \texttt{TRAIL} has an overall higher success rate and is able to navigate efficiently without incurring large experienced vertical acceleration (note again that vertical acceleration scales with vehicle speed and \texttt{TRAIL} has a greater average speed). 
Moreover, \texttt{TRAIL} generally exhibits more deterministic behavior. The only exception is Experiment III, where two of the three trials took a detour to avoid areas with short tree trunks and tall grass, while the third navigated directly through that region. This deviation is caused by the slight stochasticity of the network’s predictions, caused by sensor noise, which produced a small clearing in the geometric risk map for that particular run. 
In Experiment V, \texttt{TRAIL} found a clearing (paved road) on the right side and thus takes a detour.
From the observed behaviors of all three methods, we attribute \texttt{TRAIL}’s superior overall performance to two key factors: its ability to conduct trajectory optimization that considers the full path rather than relying purely on reactive planning, and its ability to modulate the speed profile using the predicted bumpiness of the terrain.

\subsection{Long Path Navigation}\label{subsec:long_path_nav}

We also conducted a long-range navigation experiment with \texttt{TRAIL} (route shown in \Cref{fig:hardware_exp_long_path}). Eight interventions were required. 
Five interventions were caused by limited traction on wet ground, a hardware-limited regime in which the platform’s stock tires could not provide sufficient friction for wet off-road terrain, leading to wheel slip even under conservative commands. 
The remaining three interventions were safety-related: two were due to small trees outside the camera’s field of view and within LiDAR blind spots introduced by the sensor-stack pillars, and one was caused by the trajectory optimizer overly smoothing consecutive tight turns intended to pass between nearby trees. 
Potential mitigations include equipping tires suitable for wet/off-road conditions, and algorithmically, re-normalizing the bumpiness signal with a short calibration run on the target surface and including wet/low-traction conditions during data collection.

\subsection{Limitations}
After conducting more tests in the area, we observed that sensor occlusion and field-of-view limitations are a main failure mode of \texttt{TRAIL}. 
Potential solutions include temporal fusion of sensor information. 

\section{Conclusion}
In this work, we proposed and evaluated an off-road navigation framework. 
The framework consists of a perception module that learns an implicit representation
of terrain properties and a planner that utilizes the gradient information
from the perception module to perform gradient-based trajectory optimization. 
The proposed approach is able to navigate effectively in challenging environments. 
One future direction is temporal fusion of the traversability prediction to enable longer range navigation and to overcome sensor occlusion.

\bibliographystyle{IEEEtran}
\bibliography{paper/references}

\end{document}